\pdfoutput=1
\documentclass[letterpaper, 10 pt, conference]{ieeeconf}  % Comment this line out if you need a4paper

\IEEEoverridecommandlockouts                              % This command is only needed if 
\let\labelindent\relax
\usepackage{graphics} % for pdf, bitmapped graphics files
\usepackage{booktabs}
\usepackage{subfigure}
\usepackage{multirow}
\usepackage{epsfig} % for postscript graphics files
\usepackage{mathptmx} % assumes new font selection scheme installed
\usepackage{times} % assumes new font selection scheme installed
\usepackage{amsmath} % assumes amsmath package installed
\usepackage{amssymb}  % assumes amsmath package installed
\usepackage{blindtext}
\usepackage{algorithmic}
\usepackage{textcomp}
\usepackage{xcolor}
\usepackage{threeparttable}
\usepackage{hyperref}
\usepackage[T1]{fontenc}
\usepackage{graphicx}
\usepackage{overpic}
\usepackage{makecell}
\usepackage{gensymb}
\usepackage{soul}
\usepackage{microtype}
\usepackage{enumitem}
\setlist{leftmargin=*}
\IEEEoverridecommandlockouts       
\title{\LARGE \bf
Flipbot: Learning Continuous Paper Flipping via Coarse-to-Fine Exteroceptive-Proprioceptive Exploration
}
\author{Chao Zhao\textsuperscript{*1}, Chunli Jiang\textsuperscript{*1}, Junhao Cai\textsuperscript{1}, Michael Yu Wang\textsuperscript{1,2}, Hongyu Yu\textsuperscript{1,2}, and Qifeng Chen\textsuperscript{1}
\thanks{\textsuperscript{*}Authors with equal contribution. }
\thanks{\textsuperscript{1}The Hong Kong University of Science and Technology, Clear Water Bay, Hong Kong.} 
\thanks{\textsuperscript{2}HKUST Shenzhen-Hong Kong Collaborative Innovation Research Institute, Futian, Shenzhen.}   
}      

\begin{document}
\maketitle

%%%%%%%%%%%%%%%%%%
% Todo list:
% video: 1. complete book 2. complete paper box 3. flip with different angles.
%%%%%%%%%%%%%%%%%%%%%%%%%%%%%%%%%%%%%%
\begin{abstract}

This paper tackles the task of singulating and grasping paper-like deformable objects. We refer to such tasks as paper-flipping. In contrast to manipulating deformable objects that lack compression strength (such as shirts and ropes), minor variations in the physical properties of the paper-like deformable objects significantly impact the results, making manipulation highly challenging. Here, we present Flipbot, a novel solution for flipping paper-like deformable objects. Flipbot allows the robot to capture object physical properties by integrating exteroceptive and proprioceptive perceptions that are indispensable for manipulating deformable objects. Furthermore, by incorporating a proposed coarse-to-fine exploration process, the system is capable of learning the optimal control parameters for effective paper-flipping through proprioceptive and exteroceptive inputs. We deploy our method on a real-world robot with a soft gripper and learn in a self-supervised manner. The resulting policy demonstrates the effectiveness of Flipbot on paper-flipping tasks with various settings beyond the reach of prior studies, including but not limited to flipping pages throughout a book and emptying paper sheets in a box. The code is available \href{https://robotll.github.io/Flipbot/}{here} : {\color{blue}{\url{https://robotll.github.io/Flipbot/}}}

% This paper presents Flipbot, a novel approach for singulating and grasping paper-like deformable objects. We refer to such tasks as paper-flipping. In contrast to manipulating deformable objects that lack compression strength (such as shirts and ropes), minor variations in the physical properties of the paper-like deformable objects significantly impact the manipulation results. Flipbot allows the robot to capture object physical properties by integrating exteroceptive and proprioceptive perceptions. We propose a self-supervised learning framework that learns to perform paper-flipping tasks through proprioceptive and exteroceptive inputs. By incorporating a coarse-to-fine exploration process, the system is capable of learning the optimal control parameters for effective paper-flipping. We deploy our method on a real-world robot with a soft gripper and demonstrate the effectiveness of Flipbot on paper-flipping tasks with various settings beyond the reach of prior works, including but not limited to flipping pages throughout a book and empty paper sheets in a box.

\end{abstract} 

\section{Introduction}

\begin{figure}[!t]
\vspace{0.25cm}
    \centering
    % \begin{overpic}[width=\linewidth,grid,tics=5]{img/fg1.jpg}
    \begin{overpic}[width=\linewidth]{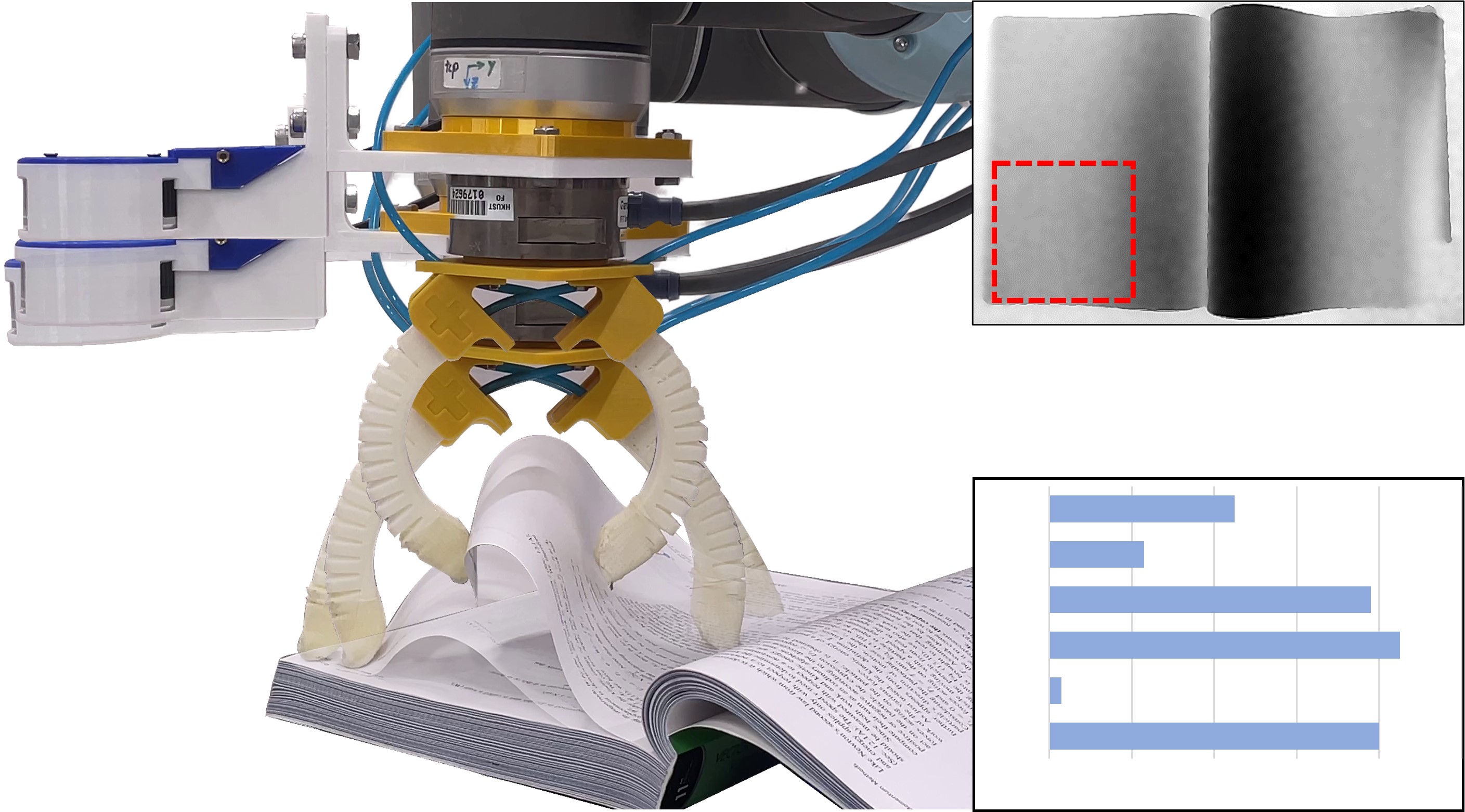}
    \put(74.75, 30.75) {\footnotesize {Exteroception}}
    \put(74.75, 1.5) {\footnotesize {Propriocepion}}
    \put(69, 35.5) {\footnotesize {Crop}}
    
    \put(67.5, 11.1) {\tiny{mx}}
    \put(67.5, 7.8) {\tiny{my}}
    \put(67.5, 5) {\tiny{mz}}
    
    \put(68.6, 14) {\tiny {fz}}
    \put(68.3, 17.5) {\tiny {fy}}
    \put(68.7, 20) {\tiny {fx}}
    \end{overpic}
    \caption{
    A soft gripper with the learned policy flips a book. The time-lapse image depicts the operation of the gripper as it interacts with the book to singulate and grasp a piece of paper. The cropped depth image in the red line box located at the upper right corner presents the exteroceptive observation from the depth camera. The readings on the bottom right show the proprioceptive observation from the force-torque sensor.
    }
    \label{fig:fg1}
\vspace{-0.5cm}
\end{figure}
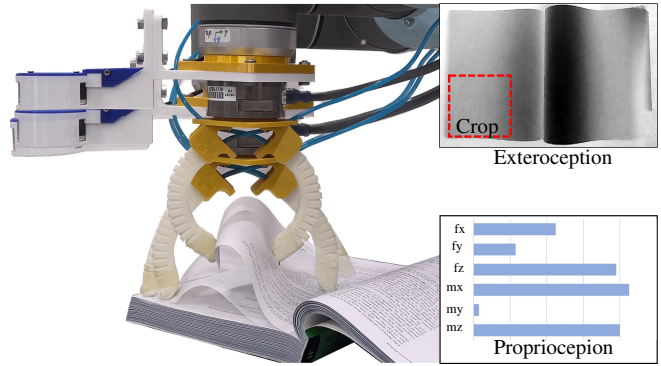

Deformable object manipulation has achieved notable progress in robotics. However, until now, robots could not match the generalization and robustness of humans in manipulating thin and flexible objects. One of these tasks is flipping book pages, as shown in Fig. \ref{fig:fg1}, which requires singulating and grasping paper page by page. Humans can briskly turn pages of a book by watching the target and using the tactile sensations on their fingertips to adjust their actions. In this process, human instinctively combines exteroceptive and proprioceptive perception to accommodate the irregular paper thickness and physical properties, such as slipperiness, stiffness, and friction. Endowing robots to have such capability is a grand challenge in the field of robotics.

One of the foremost challenges in manipulating thin and flexible objects is incomplete and noisy perception \cite{zhu2021challenges}. For example, a stack of paper is unstable, and the contact between each layer is not observable. Therefore, the robot may have to perceive physical properties between paper, such as friction, and elasticity, to successfully singulate and grasp a sheet from a stack. Exteroceptive perception obtained from camera sensors is incomplete for such tasks and unreliable in real-world conditions. The depth sensors, which most existing works rely on, cannot distinguish the different layers of stacked paper due to the paper thickness. Depth sensors are also inherently incapable of capturing the surface's physical properties, such as hardness and flexibility \cite{miki2022learning}. Some works use tactile sensors as proprioception to estimate deformable objects' physical properties. For example, \cite{yuan2016estimating} uses a high-precision tactile sensor to measure the geometry of the contact surface and the object's hardness. \cite{she2021cable} manipulates cables with a pair of robotic grippers using real-time tactile feedback. Nevertheless, high-precision tactile sensors are often expensive and require specific finger shapes to fit. In addition to the challenge in environment perception, manipulating thin and flexible objects may desire the gripper with the dexterity and compliance of human fingers, which further adds to the difficulty \cite{teeple2022multi}. 

To address the above challenges, we present Flipbot, a self-supervised method for singulating and grasping paper-like deformable objects at unprecedented robustness, enabling continuous paper flipping. At its core, Flipbot is based on a principled solution integrating exteroceptive and proprioceptive perceptions into policy learning. We obtain proprioception from the Force/Torque (F/T) sensor readings and exteroception from a depth camera. We use a procedural motion, referred to as ``Swipe'' to actively interact with the environment. When a ``Swipe'' motion is applied to a piece of paper, the deformation brought about by the interaction between the finger and object reveals imperceptible physical characteristics like mass, flexural rigidity, and friction. Meanwhile, visual observation provides global information on the environment. We design a cross-sensory encoder to integrate exteroceptive and proprioceptive perceptions into an implicit state representation. The encoder is trained end-to-end in a self-supervised manner as a part of policy learning. By incorporating exteroceptive-proprioceptive information into policy learning, the robot is able to discover the optimal policy for paper-flipping through continuous exploration. Furthermore, the reward signal for policy learning is derived from visual observation; Flipbot is fully trained by self-exploration without human demonstration or annotation. 

The primary contribution of the presented work is the proposed new approach, Flipbot, for sigulating and grasping paper-like objects. It achieves substantial improvements over the prior studies while maintaining exceptional robustness. Our extensive experiments show that Flipbot is able to perform page-flipping from the beginning to the end of a book accurately and consistently, and exhibits remarkable zero-shot generalization under conditions never encountered during training: novel paper materials such as coated and plastic paper and tasks such as emptying a box full filled with paper.

\section{Related Work}\label{sec:related}

Deformable object manipulation presents a persistent and enduring challenge within the field of robotics. Conventional analytic approaches rely on modeling object dynamics and then using model predictive control \cite{allgower2012nonlinear}, or trajectory optimization \cite{zimmermann2021dynamic} for manipulation. However, analytic approaches require substantial prior knowledge of geometry, and the physical properties of the object \cite{zhu2021challenges}. For example, \cite{flexflip} presents an approach for manipulating a piece of thin deformable object by analyzing the object's internal energy exchange concerning object poses. And \cite{guo2021deformation} proposes a close-loop shape control method utilizing visual markers, which limits the generality. Moreover, the high-dimensional state representation and complex dynamics of the deformable object provide additional challenges to generalizing novel objects and environments.

Recently, learning-based methods have become increasingly popular alternatives to perform deformable object manipulation. Most work \cite{shi2022robocraft,shen2022acid} learns the object dynamic from visual features rather than explicit modeling physical processes. For example, \cite{yan2020learning} encodes visual observation into latent space with self-supervision, followed by model-based planning. Another line of approach defines a set of primitives for deformable object manipulation and learns a mapping from image to predefined primitives \cite{ha2022flingbot}. Such image-to-primitive formulation has been applied across various tasks including manipulating rope \cite{sundaresan2020learning}, smoothing fabric \cite{avigal2022speedfolding}, and blowing bags \cite{xu2022dextairity}. However, the physical information of the environment, which necessitates deformable object manipulation, is challenging to be obtained from visual perception. In this regard, \cite{yuan2018active} estimates the physical properties of fabric materials through a high-resolution tactile sensor, GelSight \cite{yuan2017gelsight}. Further, \cite{she2021cable} proposes an approach to manipulate a cable based on tactile feedback without vision sensory. \cite{tirumala2022learning} employs tactile sensors to manually collect data for training a classifier that can differentiate between towels with thicknesses of 1-3 layers. Then a heuristic approach is used to consistently attempt to grasp specific layers of towels based on the classifier's prediction outcomes. Nevertheless, tactile sensors alone are hard to provide global information about the environment, which inevitably restricts the range of manipulation or requires prior knowledge of objects. 

More recently, a small number of papers have explored the use of soft grippers in deformable object manipulation, which is known for its ease of grasping objects without high precision control \cite{flexflip,hughes2016soft}. The authors of \cite{low2021sensorized} demonstrated a soft gripper system that is capable of handling a wide range of food products by reconfiguring fingers into different poses. In addition, \cite{teeple2022multi} quantitatively indicates that the compliance of the soft gripper can facilitate the manipulation of thin deformable objects.

Compared with the above studies, our presented approach, Flipbot, 
incorporates exteroceptive and proprioceptive feedback in deformable object manipulation rather than relying on a single perception source. Flipbot thus combines the best of both worlds: the global information about the environment afforded by exteroception and the local information about physic property afforded by proprioception. Filpbot also leverages the compliance from a soft pneumatic gripper for performing dexterous behavior. The resulting policy has taken the real robot to various tasks surpassing prior published work in the field of deformable object manipulation.

\begin{figure*}[!t]
\vspace{0.2cm}
    \centering
    % \begin{overpic}[width=\linewidth,grid,tics=5]{img/fg2.jpg}
    \begin{overpic}[width=\linewidth]{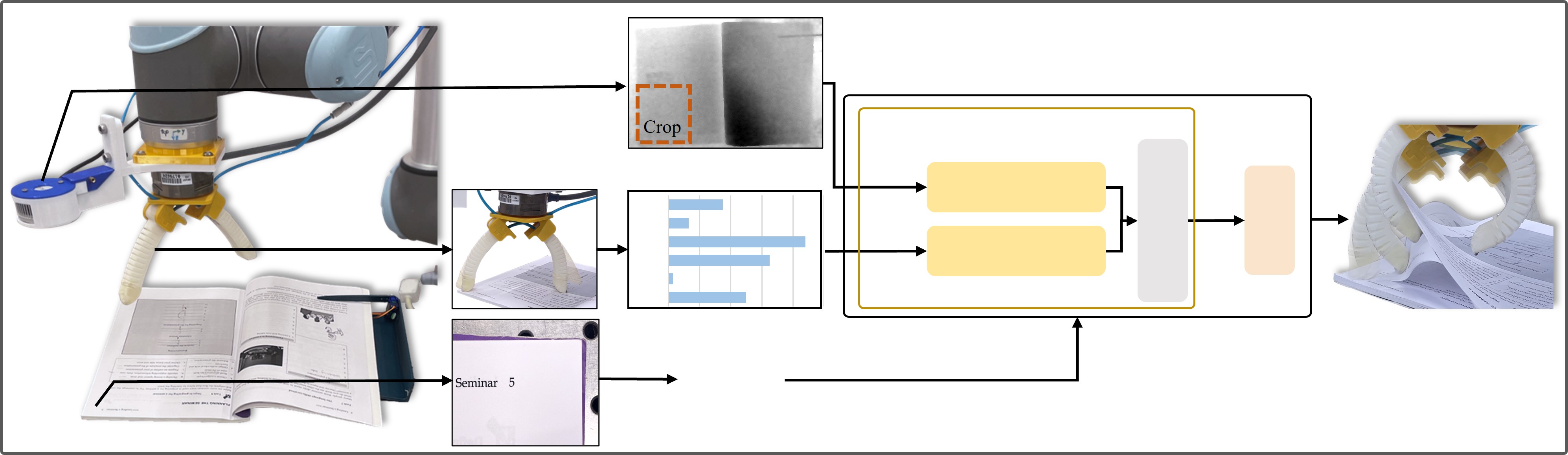}
    % \put(23,20){ {\rotatebox{90}{\scriptsize Success rate (\%)}}}
    % \put(36.5,1.7) {\small {\color{white}Depth image}}
    
    \put(41.5, 18) {\small {Exteroception}}
    \put(30.25, 9.75) {\small {``Swipe''}}
    \put(28.75, 0.85) {\small {Page number}}
    \put(7, 0.75) {\small {Real robot environment}}
    \put(41.5, 7.75) {\small {Proprioception}}
    \put(43.5, 4.25) {\small {Reward}}
    % \put(69, 6) {\small {Train with SAC}}
    \put(89, 7.5) {\small {Action $a_t$}}
    
    \put(62, 24) {\small {Policy architecture}}
    \put(57, 20.25) {\small {Cross-sensory encoder}}
    \put(59.75, 16.5) {\small {Exter encoder}}
    \put(60, 12.5) {\small {Prop encoder}}
    \put(55.75, 17.6) {\small {$o_{tv}$}}
    \put(55.75, 13.5) {\small {$o_{tf}$}}
    \put(77, 15.5) {\small {$l_t$}}
    
    \put(41, 12.2) {\tiny {mx}}
    \put(41, 11) {\tiny {my}}
    \put(41, 9.8) {\tiny {mz}}

    \put(41.4, 15.7) {\tiny {fx}}
    \put(41.4, 14.55) {\tiny {fy}}
    \put(41.4, 13.3) {\tiny {fz}}

    \put(72.7,10.5){ {\rotatebox{90}{\small Concatenate}}}
    \put(79.75,13.25){ {\rotatebox{90}{\small  MLP}}}
    \end{overpic}
    \caption{\textbf{System Overview.} 
    We train the policy using SAC in the real world. We follow a coarse-to-fine exploration process to obtain exteroception and proprioception. First, the camera captures the depth image, and the cropped area is used as extrinsic perception. Next, the soft finger ``Swipe'' on paper captures force $(fx, fy, fz)$ and torque $(mx, my, mz)$ values from force sensors as proprioception. The RL agent receives the observations and predicts the actions to be performed by the robot, and receives the reward based on changes in page numbers.
    %Operationally, exteroception and proprioception are processed separately and embedded into two latent vectors, the former using the Conv encoder and the latter using the MLP encoder. The concatenation of the two vectors is then passed through subsequent MLP layers to predict actions.
    }
    \label{fig:fg2}
\vspace{-0.5cm}
\end{figure*}
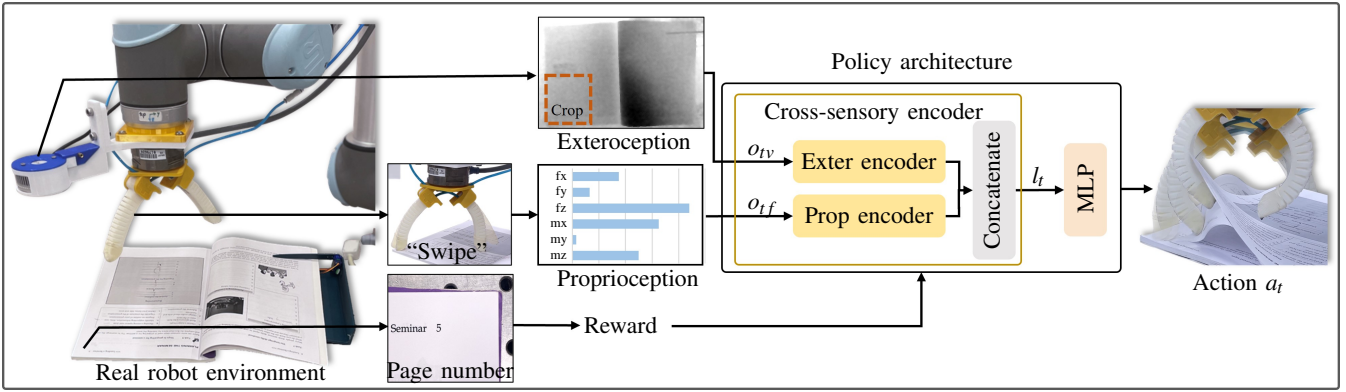

\section{Method}

The goal of Flipbot aims to empower robots to  effectively singulate and grasp thin and flexible objects through exteroceptive and proprioceptive perception. Our key insight is that global information about positions and shapes on a large scale provided by vision and local information about contact and force provided by proprioceptive perception are indispensable parts of manipulating deformable objects like paper. Also, proprioception and exteroception fusion reveals physical information that helps robots better explore and make decisions. The overview of Flipbot is shown in Fig. \ref{fig:fg2}. 

First, we utilize a simple soft gripper for manipulation (see Fig. \ref{fig:fg4}(c)). The natural compliance of the soft gripper provides unique benefits for manipulating thin and flexible objects while avoiding damage to the object. Another advantage is that the soft gripper has a more straightforward actuation strategy in movements such as bending the fingers, compared with fully actuated rigid grippers.

Then, we use a coarse-to-fine exploration process to obtain unobservable physical information about deformable objects. In this process, first, the depth camera provides a rough observation of the object. We then use a procedural motion ``Swipe'' and an F/T sensor to monitor the object's state. One advantage of using the F/T sensor instead of a tactile sensor is that the force sensor can be assembled seamlessly with soft hands without a specific finger design.

Last, we use a cross-sensory encoder to fuse the proprioception and exteroception and use model-free reinforcement learning (RL) to learn the policy that avoids explicit modeling of diverse and frequent transitions in the contact state between the object and the soft hand.

\subsection{Problem Formulation}

We formulate the problem of the paper-flipping as a Markov Decision Process (MDP). An MDP consists of four components: a state space $S$, an action space $A$, a reward function $R(s_t ,s_{t+1})$, and a transition probability $P(s_{t+1} |s_t , a_t)$. In our framework, an agent uses a policy $\pi(a_t |s_t)$ to select an action $a_t$ for controlling the robot and receives rewards $r_t$. The goal of the reinforcement learning framework is to obtain the optimal policy $\pi^*$, which maximizes the expected discounted sum of rewards over a finite time horizon. To achieve this objective, we utilize the Soft Actor-Critic \cite{haarnoja2018soft} (SAC) algorithm for training. SAC requires the learning of an actor network that maps observations to actions and a critic network that estimates the expected future rewards based on the input.

\subsection{Observations via Coarse-to-Fine Exploration}

\begin{figure}[!tp]
    \centering
    % \begin{overpic}[width=\linewidth,grid,tics=5]{img/fg3.jpg}
    \begin{overpic}[width=\linewidth]{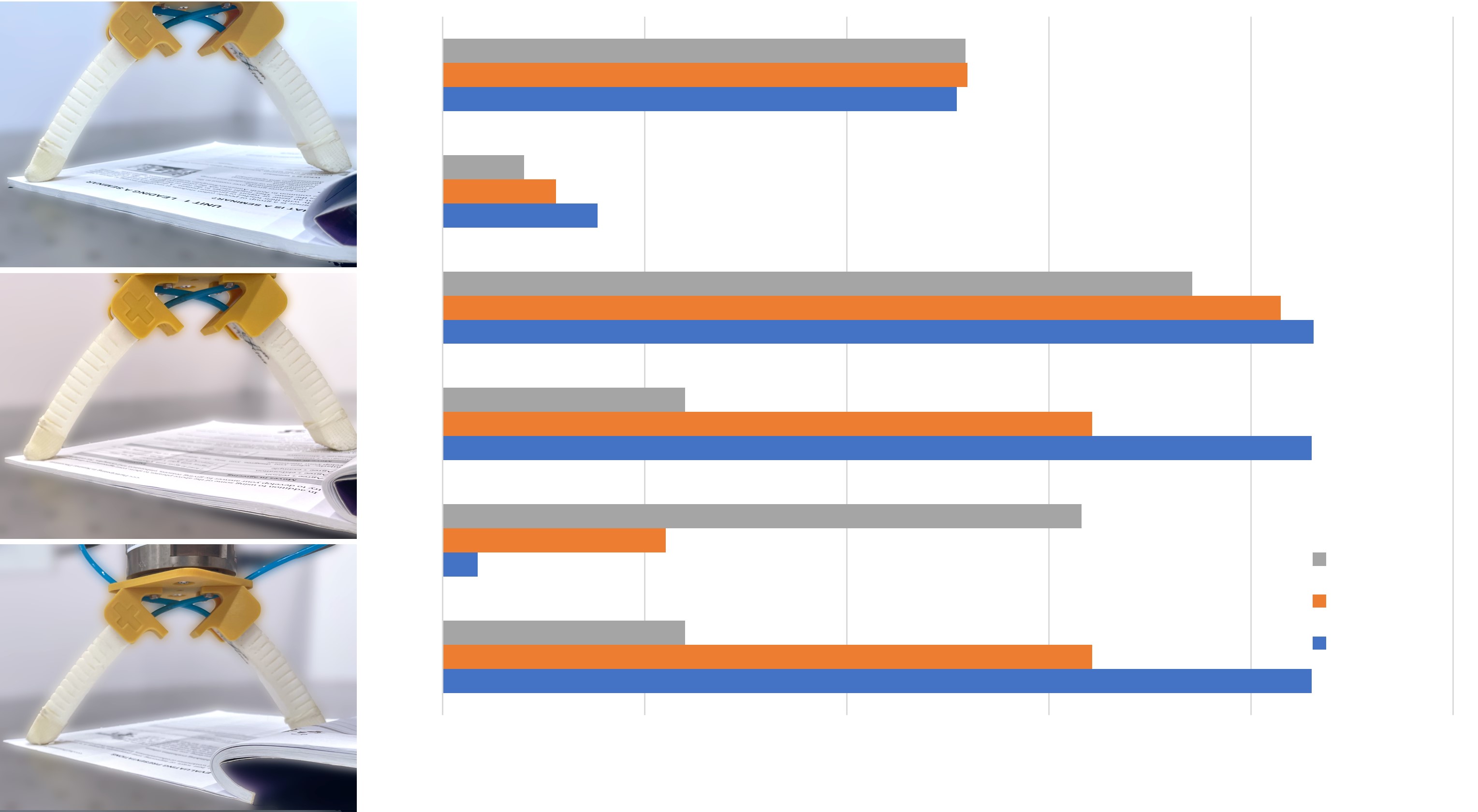}
    \put(0.75, 38.3) {\tiny{\color{white}Page 1}}
    \put(0.5, 20) {\tiny{\color{white}Page 25}}
    \put(0.5, 1) {\tiny{\color{white}Page 50}}
    \put(45, 1.5) {\scriptsize{Normalized forces and torques}}
    
    \put(26.5, 50) {\tiny {fx}}
    \put(26.5, 42) {\tiny  {fy}}
    \put(26.5, 34) {\tiny  {fz}}
    \put(26.1, 26) {\tiny  {mx}}
    \put(26.1, 18.2) {\tiny  {my}}
    \put(26.1, 10.2) {\tiny  {mz}}
    
    \put(90, 16.5) {\tiny  {Page 1}}
    \put(90, 13.55) {\tiny  {Page 25}}
    \put(90, 10.75) {\tiny  {Page 50}}
    
    \put(29.5, 5) {\tiny  {0}}
    \put(42.55, 5) {\tiny  {0.2}}
    \put(56, 5) {\tiny  {0.4}}
    \put(69.75, 5) {\tiny  {0.6}}
    \put(83.25, 5) {\tiny  {0.8}}
    \put(97.8, 5) {\tiny  {1}}
    
    \end{overpic}
    \caption{Visualization of forces and torques after ``Swipe'' on the different page numbers.
    }
    \label{fig:fg3}
\vspace{-0.5cm}
\end{figure}
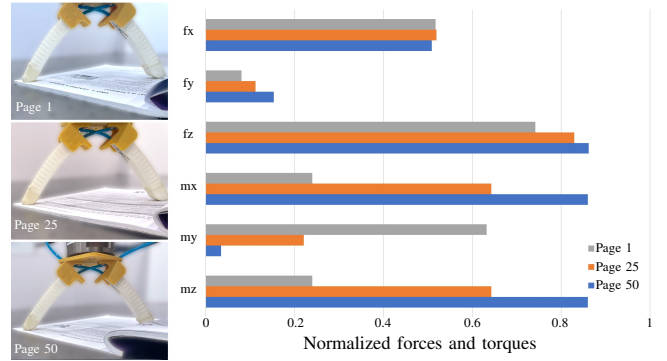

The state is defined as $s_t=(o_{tv}, o_{tf})$, where $o_{tv}$ refers to the exteroceptive observation, $o_{tf}$ refers to the proprioceptive observation, shown in Fig.~\ref{fig:fg2}. We deploy a coarse-to-fine exploration procedure with two steps for obtaining observations $o_{tv}$ and $o_{tf}$. First, a wrist-mounted camera takes the environment's point cloud $p_t$ from a height and converts the point cloud to a depth image. We then use a 60 $\times$ 60 resolution window to crop the depth image, as the exteroceptive observation $o_{tv}$. Next, we perform an exploratory ``Swipe'' motion, to obtain physical information about the contact surface between the paper and the finger. The robot first descends a certain distance that the finger of a soft hand approaches the surface of the top right corner of the paper diagonally, where the distance is calculated according to the point cloud $p_t$. Then, we give the soft gripper a positive air pressure so that fingers touch and interact with the paper. After this process, we record readings from the F/T sensor, including forces $(fx, fy, fz)$ in $x, y, z$ axes and three simultaneous torques $(mx, my ,mz)$ about the same axes. Thus, the proprioceptive observation $o_{tf}$ is defined as a tuple of $(fx, fy, fz, mx, my, mz)$. Fig. \ref{fig:fg3} shows forces and torques after ``Swipe'' on different pages in the book. By incorporating an F/T sensor and exploratory action, $o_{tf}$ latently contains rich information related to contact states between the fingers and the object, such as gripper-object friction.

\subsection{Action and Reward}
\begin{figure}[!t]
% \vspace{0.25cm}
    \centering
    % \begin{overpic}[width=\linewidth,grid,tics=5]{img/fg4.jpg}
    \begin{overpic}[width=\linewidth]{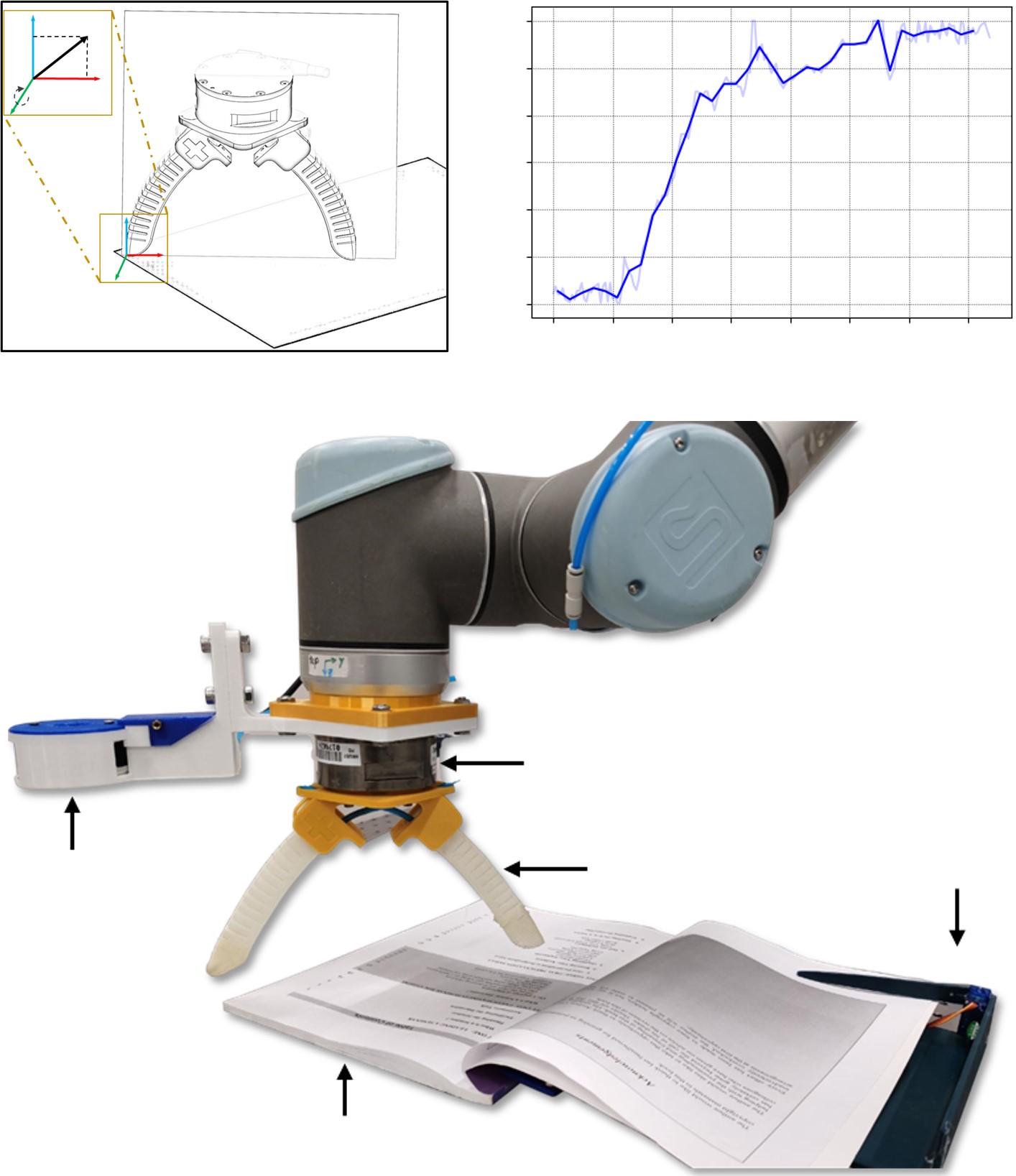}
    \put(25,96) {\scriptsize {Plane A}}
    \put(14, 77.5) {\scriptsize {$\alpha$}}
    \put(2.5, 91) {\tiny {$\theta_t$}}
    \put(5, 97.5) {\tiny {$x_t$}}
    \put(8, 95) {\tiny {$z_t$}}
    
    \put(9, 80) {\scriptsize {$\gamma$}}
    \put(10, 75) {\scriptsize {$\beta$}}
    \put(16.2,65.5) {\scriptsize {(a)}}
    \put(21,73) {\scriptsize {Paper}}
    
    \put(63.45,65.5) {\scriptsize {(b)}}
    \put(57.75,68) {\scriptsize {Training hours}}
    \put(38, 80){ {\rotatebox{90}{\scriptsize Success rate}}}
    \put(40.25, 97) {\scriptsize {100}}
    \put(41.25, 89) {\scriptsize {80}}
    \put(41.25, 81) {\scriptsize {60}}
    \put(41.25, 73) {\scriptsize {40}}
    
    \put(46, 70.25) {\scriptsize {0}}
    \put(56.3, 70.25) {\scriptsize {1}}
    \put(66.3, 70.25) {\scriptsize {2}}
    \put(76.3, 70.25) {\scriptsize {3}}

    \put(40 ,0) {\scriptsize {(c)}}
    
    \put(0, 25) {\small {Depth camera}}
    \put(16, 2.75) {\small {Book for policy training}}
    \put(50, 25.25) {\small {Soft gripper}}
    \put(45, 34) {\small {F/T sensor}}
    \put(73.5, 28.25) {\small {Recycling}}
    \put(73, 24.75) {\small {mechanism}}

    \end{overpic}
    \caption{(a): Visualization of our action coordinate system.  (b): Success rate curve of our policy training. (c): Our hardware setting for policy training.}
    \label{fig:fg4}
\vspace{-0.5cm}
\end{figure}
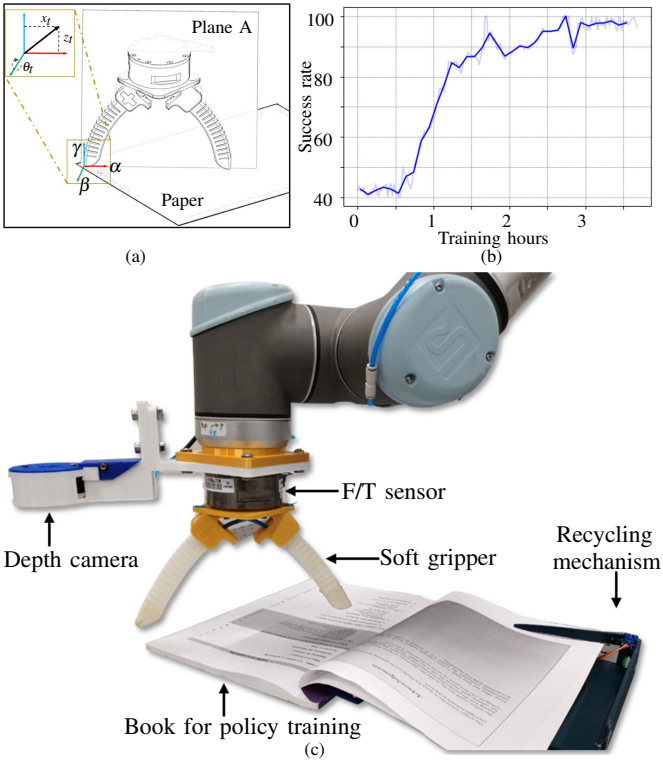

After the coarse-to-fine exploration procedure, the robot predicts the action based on observations to singulate and grasp the paper. The action includes a gripper displacement, denote as $(x_t, z_t, \theta_t)$, as shown in Fig. \ref{fig:fg4}(a). The gripper displacement refers to the relative difference between the current pose after the ``Swipe'' exploration procedure and the desired one. Specifically, $x_t \in [-6mm, 6mm]$ is the relative displacement on the line $\alpha$ connecting the two fingertips, where $\alpha$ belongs to the longitudinal plane $A$ formed by two fingers. $\theta_t  \in [0 \degree, 3 \degree]$ is the orientation of the gripper about the normal $\beta \perp A$. $z_t  \in [-6mm, 6mm]$ is the the relative displacement on the line $\gamma$, where $\gamma \perp (\alpha \times \beta)$. Furthermore, an additional action component $\Lambda$ is utilized to govern the closing or opening of the gripper. Operationally, we control the gripper aperture by commanding the pressure change. Thus, the action is formally defined as $a_t = (x_t, z_t, \theta_t, \Lambda)$, where each coordinate of the action is discretized based on the characteristics of the workspace.

At the end of an episode, the reward is given, 1 for successfully flipping a single layer of paper and 0 for otherwise. In other words, flipping two or more layers of paper simultaneously is treated as a failure. The reward is automatically determined by identifying page numbers on the book, which we describe further in Sec. \ref{sec:train_in_real}

\subsection{Policy architecture}

The policy $\pi(a_t |s_t)$ is modeled with a cross-sensory encoder and a multilayer perceptron (MLP) block, as shown in Fig. \ref{fig:fg2}. The cross-sensory encoder takes the exteroceptive observation $o_{tv}$ and proprioceptive  observation $o_{tf}$ as inputs and embeds them into a latent vector, which represents the abstraction of proprioception and exteroception. More specifically, $o_{tv}$ is processed by a global pooling layer and concatenated with $o_{tf}$ to be a vector of size 1x7. Then, the concatenated vector is fed into subsequent an MLP block to compress inputs to a more compact representation $l_t$. At last, the $l_t$ is fed through the subsequent MLP layer to predict actions.

\subsection{Training via self-supervision}\label{sec:train_in_real}
We train the policy in a real robot platform. Fig. \ref{fig:fg4}(c) shows our hardware setting for training, including the following major components: a Universal Robot 10 robot arm equipped with a 3D printed thermoplastic polyurethane soft gripper, an ATI gamma F/T sensor, and an Intel Realsense L515 depth camera, as well as a recycling mechanism. During the whole training, we only use a book assembled with printer paper, as shown in Fig. \ref{fig:fg4}(c). We train our model through trial-and-error with the following procedure: 

At each training step, the robot starts to execute coarse-to-fine exploration from an initial pose. In the process of ``Swipe'', the wrist-mounted camera captures an RGB-D image. The depth channel is used to construct exteroceptive observation $o_{tv}$, and the page number $n_t$ is recognized from RGB channels for reward calculation. The robot then descends a certain distance that the finger approaches the paper's surface to perform an exploratory action to obtain the physical observation $o_{tf}$ from the readings of F/T sensors. After this, the robot downloads the latest policy parameters from the optimizer to predict action $a_t$ and executes. We automatically calculate rewards according to the change of page numbers without human intervention, the reward $r_t$ is 1 if $n_{t+1} = n_{t} + 2$, otherwise 0. The page number identification benefits from Tesseract \cite{smith2007overview}.  At last, the generated episode is added to a replay buffer, and the optimizer sampling from this replay buffer to update the policy. We use the Adam optimizer with a learning rate of $3 \times 10^{-3}$. The robot then continuously collects episodes until it reaches the last page of the book, at which point the book is reset to the first page again using the recycling mechanism. In this way, human intervention is kept at a minimum during the training process.

The final model training took four hours, with the learning curves for the training presented in Fig. \ref{fig:fg4}(b). %The training was done on a Dell G3 laptop with an Intel Core i5-9300H processor.  

\begin{figure}[!t]
% \vspace{0.25cm}
    \centering
    % \begin{overpic}[width=\linewidth,grid,tics=5]{img/fg5.jpg}
    \begin{overpic}[width=\linewidth]{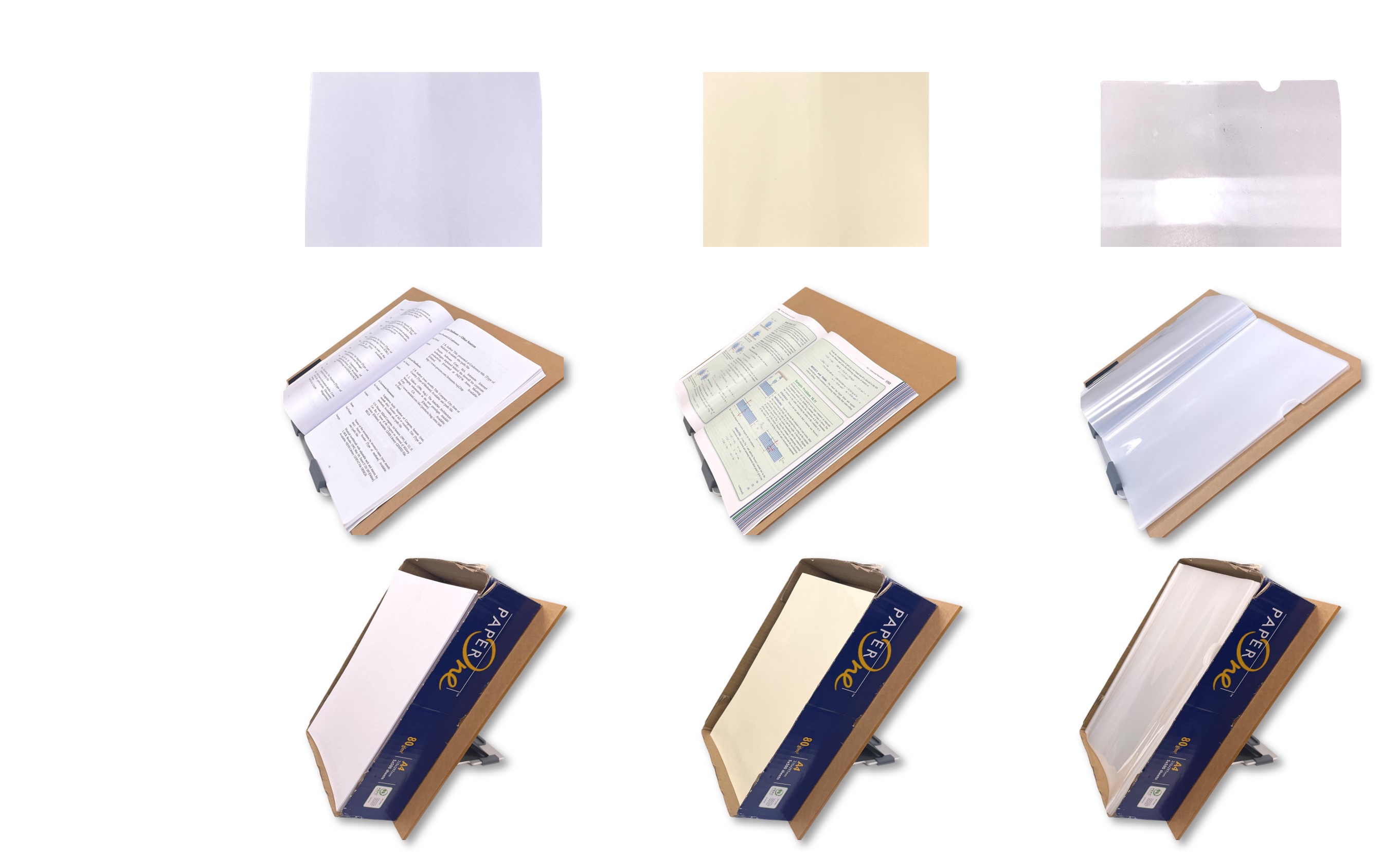}
    \put(22.5,58) {\scriptsize {Printer paper}}
    \put(50.5,58 ) {\scriptsize {Coated paper}}
    \put(79.5,58) {\scriptsize {Plastic paper}}
    
    \put(0,49) {\scriptsize {Single layer (0$\degree$)}}
    \put(3,31) {\scriptsize {Book (30$\degree$)}}
    \put(4,10) {\scriptsize {Box (60$\degree$)}}
    \end{overpic}
    \caption{A subset (9 of 27) of our test scene settings. Columns from left to right show different paper materials: printing paper, coated paper, and plastic paper. Rows from top to bottom show different test scenarios and workspace tilt angles}
    \label{fig:fg5}
\vspace{-0.5cm}
\end{figure}
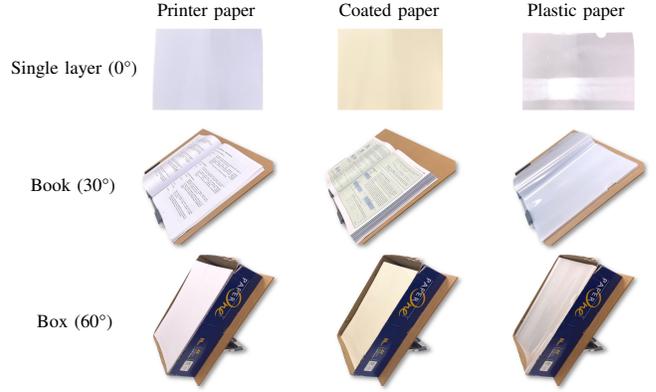

\begin{table*}[!t]
\centering
\caption{Results of experiments in the real world.}
\label{tab:real}
\resizebox{\textwidth}{!}{%
\begin{threeparttable}
\begin{tabular}{cccccccccccccccccccc} 
\hline 
\multirow{4}{*}{\begin{tabular}[c]{@{}c@{}}\\~\\Method\end{tabular}} & \multirow{4}{*}{\begin{tabular}[c]{@{}c@{}}\\~\\Tilt angle\end{tabular}} & \multicolumn{6}{c}{Full Book page flipping} & \multicolumn{6}{c}{Paper-box emptying} & \multicolumn{6}{c}{Single paper grasping} \\
 &  & \multicolumn{6}{c}{\includegraphics[scale=0.18]{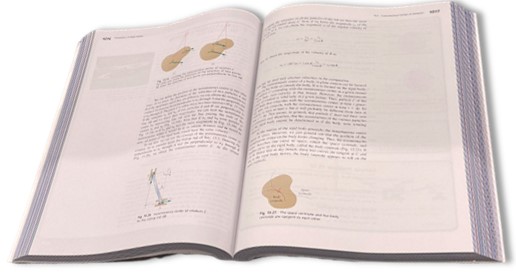}} & \multicolumn{6}{c}{\includegraphics[scale=0.18]{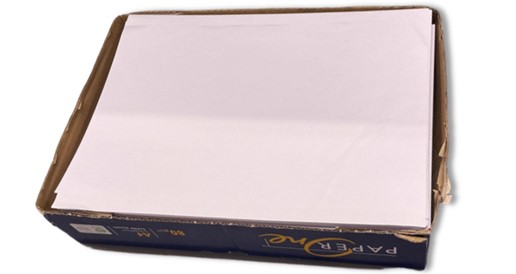}} & \multicolumn{6}{c}{\includegraphics[scale=0.18]{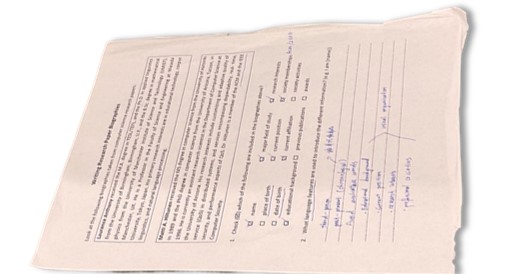}} \\
 &  & \multicolumn{2}{c}{Printer Paper} & \multicolumn{2}{c}{Coated Paper} & \multicolumn{2}{c}{Plastic Paper} & \multicolumn{2}{c}{Printer Paper} & \multicolumn{2}{c}{Coated Paper} & \multicolumn{2}{c}{Plastic Paper} & \multicolumn{2}{c}{Printer Paper} & \multicolumn{2}{c}{Coated Paper} & \multicolumn{2}{c}{Plastic Paper} \\
 &  & SR & PPH & SR & PPH & SR & PPH & SR & PPH & SR & PPH & SR & PPH & SR & PPH & SR & PPH & SR & PPH \\
\hline
Flex\&Flip \cite{flexflip} & \multirow{3}{*}{\begin{tabular}[c]{@{}c@{}}\\0$^{\circ}$\end{tabular}} & 72\% &223  & 77\% &239  & 52\% &161  & 69\% &214  & 82\% &254  & 49\% &152  & 83\% &260  & 91\% &282  & 74\% &229  \\[1ex]
Flipbot-w/o prop &  & 85\% &264  & 93\% &288  & 66\% &205  & 81\% &251  & 91\% &282  & 60\% &186  & 95\% &295  & \textbf{98\%} & \textbf{304}  & 85\% &264  \\[1ex]
\textbf{Flipbot} &  & \textbf{94\%} & \textbf{291} & \textbf{96\%} & \textbf{298} & \textbf{82\%} & \textbf{254} & \textbf{90\%} & \textbf{279} & \textbf{94\%} & \textbf{291} & \textbf{68\%} & \textbf{211} & \textbf{99\%} & \textbf{307} & \textbf{98\%} & \textbf{304} & \textbf{92\%} & \textbf{285} \\ [1ex]
\hline
Flex\&Flip \cite{flexflip} & \multirow{3}{*}{\begin{tabular}[c]{@{}c@{}}\\30$^{\circ}$\end{tabular}} &76\%  &236  &74\%  &229  &44\%  &136  &62\%  &192  &72\%  &223  &42\%  &130  &80\%  &248  &87\%  &270  &76\%  &236  \\[1ex]
Flipbot-w/o prop & &88\%  &273  &87\%  &270  &63\%  &195  &84\%  &260  &88\%  &273  &55\%  &171  &85\%  &264  &92\%  &295  &86\%  &267    \\[1ex]
\textbf{Flipbot } &  & \textbf{93\%} & \textbf{288} & \textbf{91\%} & \textbf{282} & \textbf{72\%} & \textbf{223} & \textbf{88\%} & \textbf{273} & \textbf{91\%} & \textbf{282} & \textbf{62\%} & \textbf{192} & \textbf{92\%} & \textbf{285} & \textbf{95\%} & \textbf{295} & \textbf{90\%} & \textbf{279} \\[1ex]
\hline
Flex\&Flip \cite{flexflip} & \multirow{3}{*}{\begin{tabular}[c]{@{}c@{}}\\60$^{\circ}$\end{tabular}}  &64\%  &198  &56\%  &174  &47\%  &192  &56\%  &174  &58\%  &180  &38\%  &118  &84\%  &260  &82\%  &254  &83\% &257  \\[1ex]
Flipbot-w/o prop &  &76\%  &236  &72\%  &223  &62\%  &192  &77\%  &239  &70\%  &217  &58\%  &179  &86\%  &267  &85\%  &264  &91\%  &282  \\[1ex]
\textbf{Flipbot} &  & \textbf{84\%} & \textbf{260} & \textbf{82\%} & \textbf{253} & \textbf{70\%} & \textbf{217} & \textbf{82\%} & \textbf{254} & \textbf{80\%} & \textbf{248} & \textbf{66\%} & \textbf{205} & \textbf{96\%} & \textbf{298} & \textbf{92\%} & \textbf{285} & \textbf{94\%} & \textbf{291}  \\
\hline
\end{tabular}
\begin{tablenotes}
\item[*] SR stands for success rate. 
\end{tablenotes}
\end{threeparttable}
}
\end{table*}

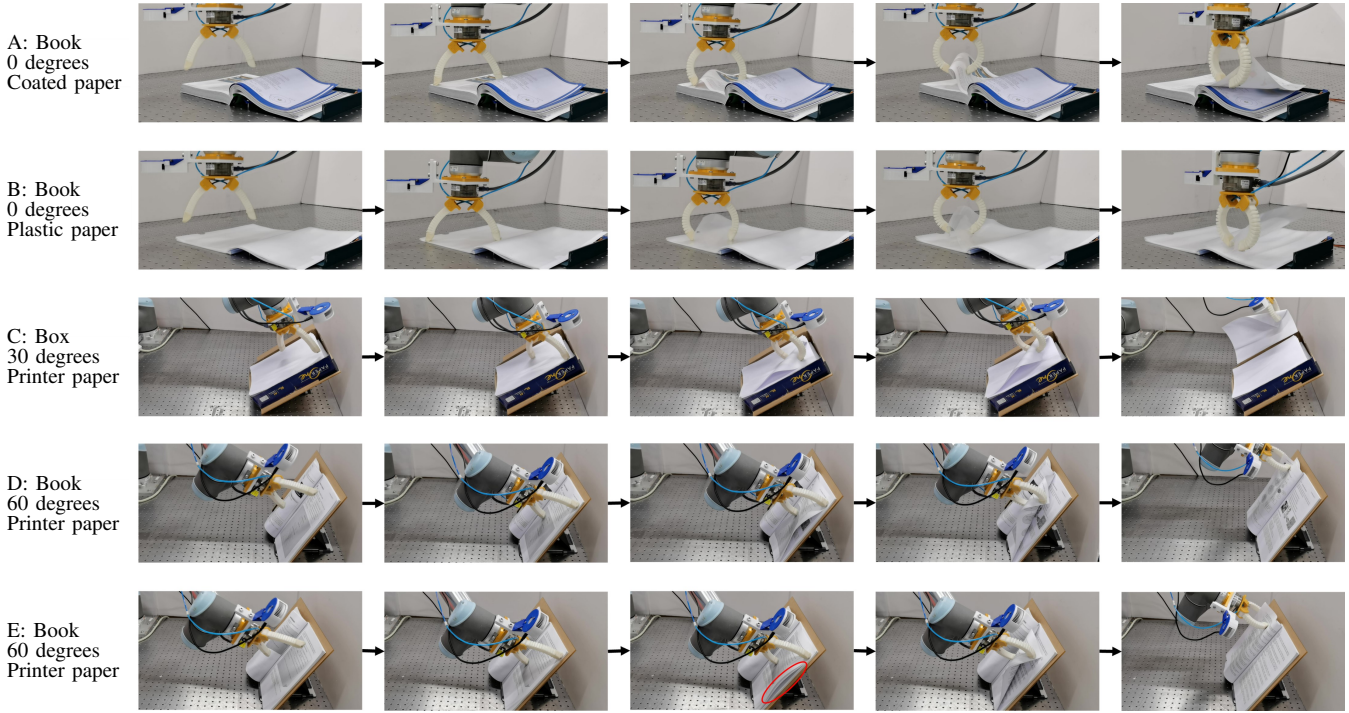
\begin{figure*}[!ht]
\vspace{0.25cm}
    \centering
    % \begin{overpic}[width=\linewidth,grid,tics=5]{img/fg6.jpg}
    \begin{overpic}[width=\linewidth]{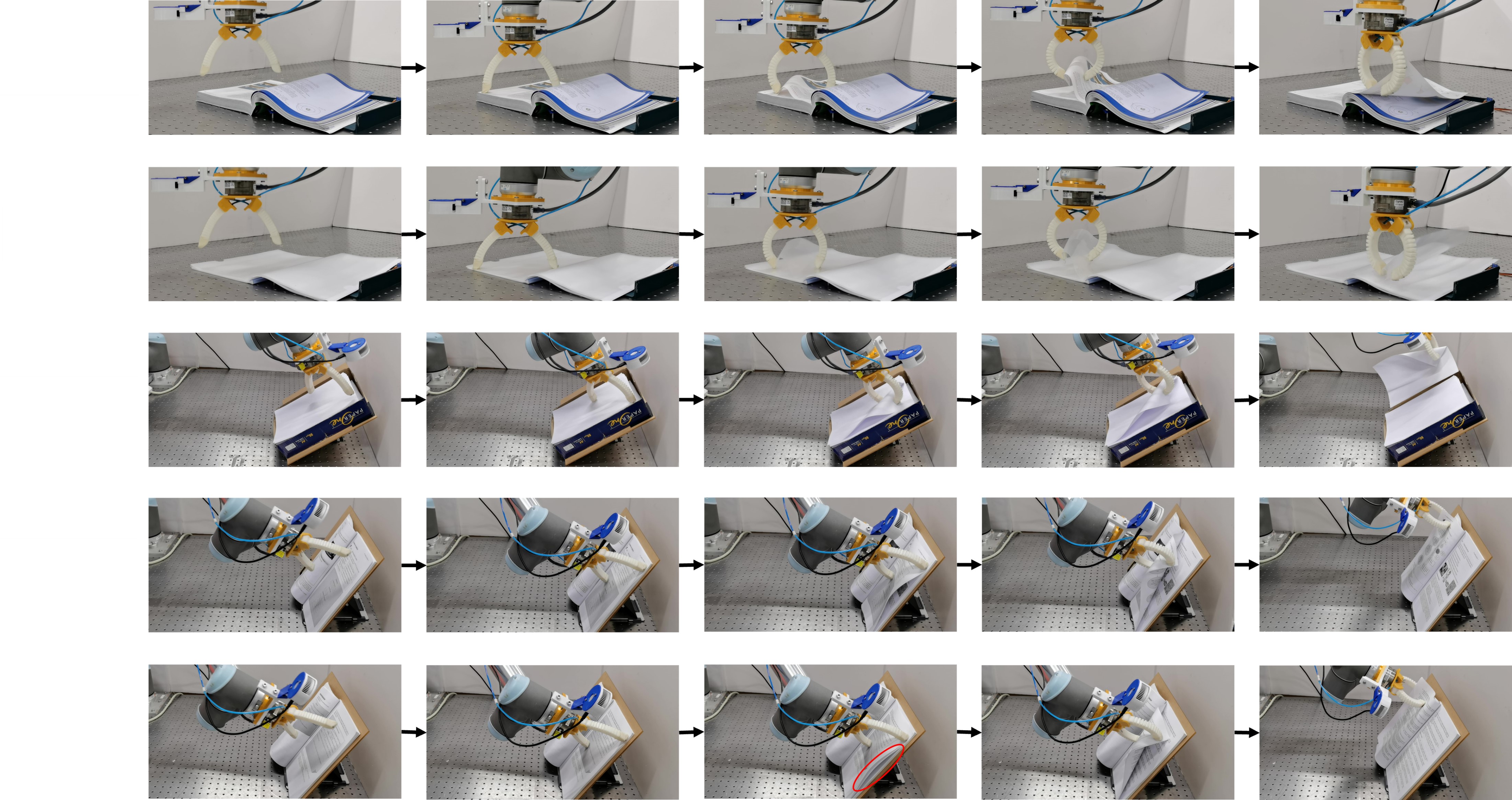}
    
    \put(0,49.5) {\footnotesize {A: Book}}
    \put(0,48) {\footnotesize {0 degrees}}
    \put(0,46.5) {\footnotesize {Coated paper}}
    
    \put(0,38.5) {\footnotesize {B: Book}}
    \put(0,37) {\footnotesize {0 degrees}}
    \put(0,35.5) {\footnotesize {Plastic paper}}

    \put(0,27.5) {\footnotesize {C: Box}}
    \put(0,26) {\footnotesize {30 degrees}}
    \put(0,24.5) {\footnotesize {Printer paper}}

    \put(0,16.5) {\footnotesize {D: Book}}
    \put(0,15) {\footnotesize {60 degrees}}
    \put(0,13.5) {\footnotesize {Printer paper}}

    \put(0,5.5) {\footnotesize {E: Book}}
    \put(0,4) {\footnotesize {60 degrees}}
    \put(0,2.5) {\footnotesize {Printer paper}}
    
    \end{overpic}
    \caption{
    Flipbot performs paper-flipping in different scenes. 
    A-D: Flipbot successfully singulates and grasps a piece of paper in various settings;
    E: Flipbot fails to singulate and grasp a piece of printer paper with a 60-degree tilt angle. The circled area in red denotes that two layers of paper were flipped.
    }
\vspace{-0.25cm}
\label{fig:fg6}
\end{figure*}

\section{Experiments}

We design a set of experiments in real-world settings to evaluate the system’s generalization ability to novel object physical parameters and the advantage of using exteroceptive and proprioceptive exploration. For all following experiments, we use the same robot hardware setting and the same model trained with the book assembled from printer paper, described in Sec. \ref{sec:train_in_real}. The system’s performance is evaluated on its generalization to unseen paper types (i.e.,  flipping different types of paper when only trained on printer paper) and unseen scenarios(e.g., emptying paper in a box) and its efficiency (i.e.,  the speed and accuracy of paper-flipping).

\textbf{Scene setup:} We investigate the performance of our system across various object settings and scene configurations. In total, we have 27 different test scenes with the combination of test scenarios, paper types and tilt angles. We test with the following three scenarios: 
\begin{itemize}
\item Full Book page flipping. It is a similar scenario as in policy training, where the robot needs to flip book pages one by one throughout the book. 
\item Paper-box emptying. The robot grasps each sheet one by one from a pile of paper dumped into a box until emptying it. This is more challenging than the book setup because the physical interaction between the paper is more complex without the constraints of the spine.
\item Single paper grasping. The robot grasps a single piece of paper lying on a flat surface. 
\end{itemize}

In each scenario, we use three types of paper that have different physical properties, including the printer paper, coated paper, and plastic paper. The physical property of printer paper is the same as we have used during training, which has the highest friction coefficient among the three types. The coated paper and plastic paper are unseen paper types. The coated paper has the lowest friction coefficient and the plastic paper has medium friction coefficient. The detailed physical properties of these three paper types are shown in Tab. \ref{tab:object prop detail}. Meanwhile, we also vary tilt angles (0, 30, 60 degrees) for the workspace to test the effect of gravity on paper flipping. 

\begin{table}
\centering
\caption{physical properties of test paper}
\label{tab:object prop detail}
\resizebox{\linewidth}{!}{% 
\begin{tabular}{cccc} 
\hline
\multirow{2}{*}{Physical properties} & Printer paper & Coated paper & Plastic paper \\
 & (seen type) & (unssen type) & (unssen type) \\ 
\hline
\begin{tabular}[c]{@{}c@{}}Static Coefficient\\ of Friction \end{tabular} & \textbf{0.462$\pm$0.0087} & 0.283$\pm$0.0104 & 0.334$\pm$0.0066    \\
\hline
\begin{tabular}[c]{@{}c@{}}Kinetic Coefficient\\ of Friction \end{tabular} & \textbf{0.417$\pm$0.0542} & 0.174$\pm$0.0229 & 0.259$\pm$0.0263   \\
\hline
\begin{tabular}[c]{@{}c@{}}Young's Modulus\\in Machine Direction($GPa$)\end{tabular} & \textbf{2.84$\pm$0.17} & 2.62$\pm$0.14 & 1.54$\pm$0.23 \\
\hline
\begin{tabular}[c]{@{}c@{}}Density\\($g/m^2$)\end{tabular} & 102.5$\pm$2.32 & 59.8$\pm$0.93 & \textbf{385.4$\pm$1.74} \\
\hline
\begin{tabular}[c]{@{}c@{}}Thickness\\($mm$)\end{tabular} & 0.096$\pm$0.006 & 0.057$\pm$0.012 & \textbf{0.151$\pm$0.017} \\
\hline
\end{tabular}
}
\vspace{-0.25cm}
\end{table}

\textbf{Metric:} We utilize two evaluation metrics for validating algorithm performance: success rates (successful paper flips/total attempts) and PPH (successful paper flips per hour). The success of paper flipping for each attempt is measured by whether the gripper detaches and flips strictly one piece of paper. For example, in the book page flipping task, the robot detaches and flips two pieces of paper simultaneously is considered a failure. PPH is the product of the speed of flipping in an hour and the success rate, which includes the time of perception, network inference, and robot execution in enabling paper-flipping manipulation. It is important to note that our Flipbot implementation is not optimized for high-speed execution; thus, the reported PPH is solely used to compare relative performance.

% \noindent 
\textbf{Algorithm comparisons:} We compare with the following methods:
\begin{itemize}
\item \textbf{Flex\&Flip \cite{flexflip}:} it simplifies a piece of paper as a linear object and uses a physical model to analyze the motion. Its original version could only grasp a single piece of paper lying on a flat surface. We adapt and extend the physical model provided by the authors and hardcode the thickness of different paper types to allow for multi-layered paper flipping.
\item \textbf{Flipbot-w/o prop:} policy learns from only exteroceptive sensory (i.e., depth camera), which directly maps the visual observation to action.
\item \textbf{Flipbot:}  policy learns with coarse-to-fine exteroceptive-proprioceptive exploration, which is the full non-ablated method we propose in this article.
\end{itemize}

\subsection{Experimental Results}

\textbf{Comparison to prior work.} We first compare the performance of our approach with Flex\&Flip \cite{flexflip} with different paper types and scenarios (row 1 vs. row 3 in Tab. \ref{tab:real}). Note that Flex\&Flip \cite{flexflip} is the state-of-the-art method for single-layer paper grasping, and we extend it to multi-layered paper scenarios (i.e., paper-box emptying and full book page flipping). In the single paper grasping case, Flipbot performs better (+16\%) than Flex\&Flip \cite{flexflip} on printer paper. The advantage is much more pronounced in multi-layered paper cases, with Flipbot outperforming Flex\&Flip \cite{flexflip} 
around 20\%. In all three test scenarios, quantitative results in Tab. \ref{tab:real} suggest that our method (Flipbot) maintains comparable success rates on unseen paper types (i.e., coated and plastic paper)  with respect to the seen paper type (i.e., printer paper). In contrast, the performance of Flex\&Flip \cite{flexflip} on the plastic paper type degrades significantly (up to -20\%) on unseen paper types.

\textbf{Effectiveness of exteroceptive-proprioceptive exploration.} We conduct controlled experiments to evaluate the contribution of exteroceptive-proprioceptive exploration quantitatively. The proprioceptive perception provides information on the unobservable physical features,  facilitating policy learning effectiveness. As a result, compared with Flipbot-w/o prop that does not use proprioceptive, Flipbot achieved a higher success rate. Quantitative results in Tab. \ref{tab:real} indicate that compared to Flipbot-w/o prop, the success rate of Flipbot increases at most 24\% and at least 4\% across test cases.

\textbf{Generalization to novel tilt angles of workspace.}
In this experiment, we investigate the generalization ability of these methods to gravity changes by varying tilt angles (0, 30, 60 degrees) of the workspace (see Fig. \ref{fig:fg6}C-D). In different tilt angle setups, detaching a single sheet of paper becomes more challenging as the physical properties between the different layers of the paper change with the direction of gravity. Quantitative results in Tab. \ref{tab:real} show that the performance of our learned policy degrades slightly as the tilt angle increases. We hypothesize this happened since the physics in these test scenes differ from the training, increasing the difficulty of generalization. Nevertheless, Flipbot still outperforms other methods in terms of success rate and PPH in all test cases.

Overall, our experimental evaluation demonstrates that Flipbot is an efficient approach for paper-flipping tasks. We find the exteroceptive and proprioceptive perceptions are essential for paper-flipping, particularly for sigulating and detaching a sheet from a pile of paper. The learned policy has been demonstrated to outperform state-of-the-art methods and is also applicable to tasks beyond the reach of prior studies, such as turning pages throughout a book. Our work is not without limitations. First, when the working area is at a larger inclination angle, the friction between the paper tends to be smaller. Hence, multiple layers of paper are easy to be grasped simultaneously (see Fig. \ref{fig:fg6}E). Also, two layers of paper sometimes stick together. We assume it happens because of Van der Waals forces. A dual-arm system may be essential to address this issue, suggesting exciting opportunities for future study.

\section{Conclusion}

We have presented a novel solution for singulating and grasping thin and flexible deformable objects that utilize the cross-sensory encoding of exteroceptive and proprioceptive perceptions, which we term Flipbot. Meanwhile, the system takes advantage of the under actuation and compliance of the soft pneumatic actuator to control contact forces precisely for the singulation of a thin layer of deformable objects. We deploy the algorithm on a real-robot system and show that integrating exteroceptive and proprioceptive inputs can effectively facilitate deformable object manipulation. Extensive controlled experiments demonstrated the robustness and effectiveness of Flipbot. Beyond the experiment results, our work extends frontiers in deformable object manipulation, and the methodology presented in this work can have broad applications. A future direction is to extend the proposed approach to long-horizon deformable object manipulation tasks, such as origami folding, cleaning messy desktops, collecting mail and letters, etc.

% The method of using soft gripper for soft material manipulation may have huge potential in dealing with domestic daily tasks without damaging to the objects and harm to humans. For instance, helping disabled people to read books, clean messy desktop with documents and collect mails and letters, etc.
% 

%%%%%%%%%%%%%%%%%%%%%%%%%%%%%%%%%%%%%%%%%%%%%%%%%%%%%%%%%%%%%%%%%%%%%%%%%%%%%%%%

\bibliographystyle{IEEEtran} % use IEEEtran.bst style
% \bibliography{}
\bibliography{IEEEabrv,references}

\end{document}